\documentclass{article}

\PassOptionsToPackage{numbers}{natbib}

\usepackage[final]{neurips_2020}

\usepackage[utf8]{inputenc} % allow utf-8 input
\usepackage[T1]{fontenc}    % use 8-bit T1 fonts
\usepackage{hyperref}       % hyperlinks
\usepackage{amssymb}
\usepackage{url}            % simple URL typesetting
\usepackage{booktabs}       % professional-quality tables
\usepackage{amsfonts}       % blackboard math symbols
\usepackage{nicefrac}       % compact symbols for 1/2, etc.
\usepackage{microtype}      % microtypography
\usepackage{xcolor}         % colors
\usepackage{times}
\usepackage{amsmath, amssymb}
\usepackage{mathtools}
\usepackage{latexsym}
\usepackage{todonotes}
\usepackage{comment}
\usepackage{tabto}
\usepackage{soul}
\usepackage{multicol,multirow}
\usepackage{graphicx}
\usepackage{verbatim}
\usepackage{authblk}

\def\BibTeX{{\rm B\kern-.05em{\sc i\kern-.025em b}\kern-.08em
    T\kern-.1667em\lower.7ex\hbox{E}\kern-.125emX}}
    
\usepackage{color, colortbl}    
\usepackage{amssymb}
\usepackage{algorithm}
\usepackage[noend]{algpseudocode}
\usepackage{xcolor}

\title{Federated Continual Learning through distillation in pervasive computing }

\author[1]{\textbf{Anastasiia Usmanova}}
\author[2]{\textbf{François~Portet}}
\author[2]{\textbf{Philippe Lalanda}}
\author[2]{\textbf{German Vega}}

\affil[1]{Univ. Grenoble Alpes, France}

\affil[2]{Univ. Grenoble Alpes, CNRS, Inria, Grenoble INP, LIG, 38000 Grenoble, France  \qquad}

\begin{document}

\maketitle

\begin{abstract}

Federated Learning has been introduced as a new machine learning paradigm enhancing the use of local devices.  At a server level, FL regularly aggregates models learned locally on distributed clients to obtain a more general model. Current solutions rely on the availability of large amounts of stored data at the client side in order to fine-tune the models sent by the server. Such setting is not realistic in mobile pervasive computing where data storage must be kept low and data characteristic can change dramatically. To account for this variability, a solution is to use the data regularly collected by the client to progressively adapt the received model. But such naive approach exposes clients to the well-known problem of catastrophic forgetting. To address this problem, we have defined a Federated Continual Learning approach which is mainly based on distillation. Our approach allows a better use of resources, eliminating the need to retrain from scratch at the arrival of new data and reducing memory usage by limiting the amount of data to be stored. This proposal has been evaluated in the Human Activity Recognition (HAR) domain and has shown to effectively reduce the catastrophic forgetting effect. 
\end{abstract}
\medskip

\section{Introduction}

Pervasive computing promotes the integration of smart electronic devices in our living and working spaces in order to provide a wide variety of services. This concept, which has been around for many years now \cite{becker}, is becoming increasingly popular and is now being applied in many areas. This is due in particular to the emergence of robust and accurate smart devices, to advances in software engineering to deal with dynamic, unpredictable execution environments, and to the increasing use of machine learning (ML), which allows to develop prediction models from data collected in the field. ML techniques are particularly suitable for pervasive systems where traditional solutions cannot be used because of excessive algorithmic complexity.

\medskip

The field of wearable devices is particularly representative of those recent developments. Smartphones or smart watches, for instance, are equipped with many high quality Inertial Measuring Units (IMU) sensors that open the way to a number of personal services. In particular, wearable devices are today used  to automatically identify and monitor the basic activities of a person. Today, most solutions to recognize human basic movements are based on ML techniques that make use of the continuous streams of data provided by the worn sensors. 

\medskip

The use of machine learning-based models in pervasive systems has thus shown all its interest and is widely studied today. It allows to face evolving, uncertain, and incompletely characterized environments. However, it raises unprecedented technological and scientific challenges, notably by calling into question current software engineering practices. This includes, in particular, the definition of appropriate software architectures and the life-cycle management of the prediction models within these architectures. Most current approaches are based on architectures where prediction models are built and executed in cloud infrastructures. The results of the predictions are then applied in the field. This solution raises cost and reactivity issues and is progressively replaced by more decentralized approaches where the models are computed in the cloud but executed as close to the devices as possible, most often on an edge machine. However, such distribution of data and calculations, in turn, suffers from limitations related to security (data has to be sent over unprotected networks in order to compute new prediction models), performance (high volume of data may have to be transported), and communication costs. 

\medskip

Federated Learning (FL) \cite{mcmahan} \cite{bonawitx} is a new paradigm where models are executed and specialized on edge devices. On a regular basis, the models computed by the devices are uploaded to a cloud infrastructure to be aggregated and sent back down to the devices. This approach, where only the models cross the networks, is safer and less expensive. However, specialization of models at the device level is complex and is exposed to the well-known problem of \emph{catastrophic forgetting} \cite{ewc}, which appears when a neural network is optimized on new data that are too different from the data previously used for training.

\medskip

The purpose of this paper is to present a Federated Continual Learning (FCL) solution based on distillation, declined in two versions. These approaches enable the accuracy of the models to be maintained while effectively reducing the catastrophic forgetting effect. It has been experimented in the mobile Human Activity Recognition (HAR) domain. The paper is organized as it follows. In Section~2, we present the experimental settings that have been developed to evaluate our proposals.  Our approach is presented in detail in Section~3. Next sections are dedicated to experimental results, discussion, and conclusion.

\section{Experimental settings}

\subsection{Evaluation domain }\label{sec:har}

To evaluate our proposals, we chose the field of Human Activity Recognition with smartphones. This domain lends itself very well to this work: it is representative of the evolutions in pervasive computing, it strongly uses ML techniques and decentralization to the edge, and has strong privacy and security constraints. Also, there are several open source datasets that can be used as a basis for comparison with existing work in the community. More precisely, we have selected the UCI \cite{uci} dataset, which is commonly used in research work. It includes accelerometer and gyroscope data collected by about 30 people with a smartphone (Samsung Galaxy S II) worn on their waist. Six activities have been recorder: \textit{Walking} (activity 0), \textit{Walking Upstairs} (activity 1), \textit{Walking Downstairs} (activity 2), \textit{Sitting} (activity 3), \textit{Standing} (activity 4), and \textit{Lying} (activity 5). 

\medskip

The prediction model used in this study is a Convolution Neural Network (CNN) as defined in \cite{percom}. Precisely, this model includes 196 filters of a 16x1 convolution layer, a 1x4 max pooling layer, 1024 units in a dense layer, and a softmax layer. This model has been trained with the UCI dataset with a train/validation/test partition of 70\%, 15\% and 15\%. The learning phase was based on a mini-batch SGD of size 32 and a dropout rate of 0.5. The resulting model has an accuracy performance of 94.64\%. All experiments have been developed on top of TensorFlow 2 with Python 3, and executed on Intel(R) Xeon(R) 2.30GHz (2 CPU cores, 12GB available RAM).

\subsection{Federated learning infrastructure}

We have set up a test infrastructure to evaluate our proposals on Federated Continual Learning. The infrastructure is made of a server and several connected laptops that emulate clients. Clients are fed with data from the UCI dataset and have to predict the activity classes, from 0 to 5 as indicated earlier. At first, clients all use the same CNN model as defined here before. Then, models are updated (trained) locally and naturally diverge. 

\medskip

The server, on its side, is in charge of aggregating the clients' models as defined by the FL paradigm and, then, send them back to the clients. Aggregation is done on a regularly basis since clients send up their model synchronously, and is performed with the FedAvg algorithm, which is simple and efficient in the HAR domain \cite{percom}.

\medskip

Local training is done using sequences of disjoint tasks. A task is here defined as a set of classes in a dataset. Formally, each client $k \in \{ 1, 2, ..., K \}$  has its privately accessible sequence of $n_k$ tasks $\mathcal{T}_k$:
\vspace{-0.1cm} 
\begin{equation}
\mathcal{T}_k =  \left[\mathcal{T}_k^{1}, \mathcal{T}_k^{2}, ... , \mathcal{T}_k^{t}, ... , \mathcal{T}_k^{n_k}\right],
\mathcal{T}_k^{t} = (C^t_k, D^t_{k}),  \nonumber
\label{eq:task_seq_fcl}
\end{equation}

\noindent where $t \in \{1,...,n_k\}$, $C^t_k$ is a set of classes representing the task $t$ of a client $k$ (such that $C^{i}_k \cap C^{j}_k= \varnothing$ if $i \neq j$) and $D^t_k =  \{X^t_k,Y^t_k\}$ is training data corresponding to $C^t_k$.

\vspace{0.15cm}

A task can be carried out over several FL rounds. Formally, each task $\mathcal{T}_k^{t}$ for client $k$ is used for training during $r^t_k$ communication rounds and $\sum_{t=1}^{n_k}r^t_k = R$, where $R$ depicts the number of rounds.

\medskip

In order to emulate real life execution conditions,  data used by clients is changed at each FL round. Similarly, the tasks can evolve between each round. A client may see previously unknown tasks.  Formally, at communication round $r$ client $k$ uses training data $D_{kr} = \{X_{kr}, Y_{kr}\}$:

\begin{equation}
D_{kr} = D^t_{kr} \subset D^t_k, \hspace{2mm}  \sum_{d=1}^{t-1}r^d_k < r \leqslant \sum_{d=1}^{t-1}r^d_k + r^t_k, \nonumber
\end{equation}

\noindent  where $D_{k'r'} \cap D_{k''r''} = \emptyset $, if $k' \neq k''$ and $r' \neq r''$

\noindent ($1 \leqslant k',k'' \leqslant K, 1 \leqslant r',r'' \leqslant R$);

\subsection{Federated Learning scenario}\label{sec:exp_FL}

We then defined a scenario that favors catastrophic forgetting \cite{comorea}. This scenario is based on an observable client for which specific sequences of tasks are defined: this client is denoted \textbf{Client 1}. The scenario is based on eight FL communication rounds. For four rounds, Client 1 performs and learns a single task, that is (\emph{Walking upstairs}). In the next four, it learns another single task, that is (\emph{Walking downstairs}). Formally, it can be denoted that Client 1 learns $n_1 = 2$ tasks in total: $\mathcal{T}_1^1 = (C_1^1, D_1^1)$ and $\mathcal{T}_1^2 = (C_1^2, D_1^2)$, where $C^1_1 = \{1\}$ and $C^2_1 = \{2\}$; $\mathcal{T}_1 = \left[\mathcal{T}_1^{1}, \mathcal{T}_1^{2}\right]$,  $r_1^1=r_1^2=R/2$.  At each round, Client 1 model is initialized with the model coming from the server and subsequently updated with data collected between the communication rounds $D^t_{1r} \subset  D_1^t $.

\medskip

We also defined $K-1$ other clients in the FL architecture. These clients are similar in the sense that they perform the same well-balanced task, which includes examples of all classes at each FL round. So, they learn $n_g = 1$ task in total: $\mathcal{T}_g^1 = (C_g^1, D_g^1)$, where $C^1_g = \{0,1,2,3,4,5\}$, and have their privately accessible sequence of tasks $\mathcal{T}_g = \left[\mathcal{T}_g^{1}\right]$, so $r_g^1=R$. At each FL round, the dataset size is the same for all clients in order to measure the forgetting on equal grounds. Also, we make the assumption that the $K-1$ clients behave in the same way. Thus, their contribution to the FL process can be modeled by a single \textbf{generalized client}. This is why, for aggregation at the server side, the weights of generalized and observed clients are $1/K * (K-1)$ and $1/K$. 

\medskip

In order to implement our scenario, examples  have been randomly taken from the UCI dataset so as to form the train and test datasets. For each client $k$ and for each FL round $r$, we built a train set $D_{kr}$ of the same size. Models performance have been calculated on a same test set for all clients. The latter comprises 100 examples of each class, that is 600 in total. Eight FL rounds and ten epochs are implemented for the clients local training. We assumed that we have $K=5$ clients (the influence of $K-1$ of them represented by a single generalized client). The size of a dataset for each client $k$ and each round $r$ is $|D_{kr}| = 120$. We used a learning rate $\eta = 0.01$, dropout rate equal to 0.5, batch size $B = 32$ and SGD optimizer.

%%%%%%%%%%%%%%%%%%%%%%%%%%%%%%%%%%%%%%%%%%%%%%%%%%%%%%%%%%
%%%%%%%%%%%%%%%%%%%%%%%%%%%%%%%%%%%%%%%%%%%%%%%%%%%%%%%%%%
%%%%%%%%%%%%%%%%%%%%%%%%%%%%%%%%%%%%%%%%%%%%%%%%%%%%%%%%%%

\section{Proposal}

In order to deal with catastroohic forgetting in the FL approach, we propose to use distillation, whose original
purpose is to transfer knowledge from a large model (\emph{teacher}) to a smaller one (\emph{student}). 
More precisely, we developed two approaches:

\begin{itemize}
  \item FLwF-1 (Federated Learning without Forgetting - 1) relies on a unique teacher, that is the past model of the client (let us say at FL round $r-1$). The student is the current client model (FL round $r$). 
  \item FLwF-2 (Federated Learning without Forgetting - 2) relies on two teachers, that is the past model of the client (at round $r-1$) and the current server model. Initially, the current client model and the server model are the same. 
\end{itemize}

Using the past client model for distillation (FLwF-1) is a recent practice in CL \cite{lwf}. However, using the server model as well for distillation (FLwF-2), to the best of our knowledge is the first attempt. Testing these two models will enable to evaluate the contribution of this new distillation based on the server model.

\subsection{FLwF-1}

Our first proposal is named FLwF-1 because it defines \textbf{one} teacher (past model of the client, saved from the previous FL round) and one student (current model of the client, initially sent by the server). Our proposal is based on the computation of a distillation loss and a classification loss, which are then combined to get a final loss. Distillation loss calculates the difference between student predictions and teacher predictions, while classification loss calculates the  difference between student predictions and ground-truth labels. 

\smallskip

The output logits of the teacher model are denoted  $\mathbf{o}^{r-1}(x)=\left[o^{r-1}_{1}(x), \ldots, o^{r-1}_{n}(x)\right]$, where  $x$ is an input to the model. The output logits of the student model are denoted  $\mathbf{o}^{r}(x)=\left[o^{r}_{1}(x), \ldots, o^{r}_{n}(x)\right]$.

\smallskip
\smallskip

\textit{Distillation loss} for client $k$ and communication round $r$ is defined as follow:
\vspace{-0.2cm}
\begin{equation}
\label{eqn:dis_client}
L_{dis\_cl}(D_{kr}; \theta^{k}_{r}, \theta^{k}_{r-1}) = \sum_{x\in X_{kr}} \sum_{i=1}^{n}-\pi_{i}^{r-1}(x) \log \left[\pi_{i}^r(x)\right], \nonumber
\end{equation}
\vspace{-0.3cm}

 \noindent where $\theta^{k}_{r}$ represents the weights of the current model of client $k$ in communication round $r$, $\theta^{k}_{r-1}$ represents the weights of the previous client model,  $D_{kr} = \{X_{kr},Y_{kr}\}$ is the dataset used during  communication round $r$ by client $k$, and $\pi_{i}^{r'}(x)$ are  temperature-scaled logits of the model. They are defined as it follows:

\vspace{-0.2cm}

\begin{equation}
\pi_{i}^{r'}(x) =\frac{e^{o_{i}^{r'}(x) / T}}{\sum_{j=1}^{n} e^{o_{j}^{r'}(x) / T}}, \nonumber
\end{equation}

\noindent where $T$ is the temperature scaling parameter \cite{dist2}. Temperature-scaled logits  $\pi_{i}^{r-1}(x) $ refer to predictions of the teacher model ($\mathbf{o}^{r-1}(x)$) and $\pi_{i}^{r}(x) $ refer to predictions of the student model ($\mathbf{o}^{r}(x)$).

\medskip
\medskip
\medskip
\medskip
\medskip

\textit{Classification loss} (softmax cross-entropy) is defined as: 

\vspace{-0.4cm}
\begin{equation}
\label{eqn:loss_class}
L_{class}(D_{kr}; \theta^{k}_r) = \sum_{(x,y) \in D_{kr}} \sum_{i=1}^{n} -y_i \log \frac{exp(o^r_i(x)}{\sum_{j=1}^{n}exp(o^r_j(x))}, \nonumber
\end{equation}

\noindent where  $(x,y) \in D_{kr} = \{X_{kr}, Y_{kr}\}$ and $D_{kr} \subset D_{kr}^t$; $x$ is a vector of input features of a training sample, $y$ corresponds to some class of a set $C^t_k$ and presents as a one-hot ground truth label vector corresponding to $x$: $y\in\{0,1\}^{n=6}$.

The final loss for each client $k$ consists of a classification loss and distillation loss computed with the model of client $k$ for round $r$ and the model of client $k$ for round $r-1$: 

\vspace{-0.15cm}
\begin{equation}
L_{FLwF-1} = \alpha L_{class} + (1-\alpha)L_{dis\_cl}, \nonumber
\end{equation}

\noindent where $\alpha$ is a scalar which purpose is to regularize the influence of each term.

\subsection{FLwF-2}

Our second proposal is named  \textbf{FLwF-2} because it defines \textbf{two} teachers. It seeks to take advantage of the server model that keeps a general knowledge of all the clients. Our approach here is then to use the server model as a second teacher. The first teacher (past model of the client) can increase the specificity of the client, which allows it to perform well on tasks it has learned before. The second teacher (server model) can improve generality of the client by transferring knowledge from all the other clients and avoid over-fitting when dealing with new tasks. For the first communication round, only the server model is used as a teacher.

\smallskip

The output logits of the server model (second teacher) are denoted  $\hat{\mathbf{o}}^{r-1}(x)=\left[\hat{o}_{1}^{r-1}(x), \ldots, \hat{o}_{n}^{r-1}(x)\right]$. 

\smallskip

\textit{Distillation loss for the server model} is defined as:

\vspace{-0.5cm}
\begin{equation}
\label{eqn:dis_server}
L_{dis\_serv}(D_{kr}; \theta^{k}_{r}, \theta_{r-1}) = \sum_{x\in X_{kr}} \sum_{k=1}^{n}-\hat{\pi}_{k}^{r-1}(x) \log \left[\pi_{k}^r(x)\right], \nonumber
\end{equation}

\vspace{-0.2cm}

\noindent where  $\theta^{k}_{r}$ represents the weights of the current model of client $k$ in communication round $r$, $\theta_{r-1}$ represents the weights of the server model after communication round $r-1$, and $\hat{\pi}_{k}^{r-1}(x)$ is the temperature-scaled logits of the server network. 

The final loss for the proposed method is:

\vspace{-0.4cm}
\begin{equation}
\label{eqn:final}
L_{FLwF-2} = \alpha L_{class} + \beta L_{dis\_cl} + (1-\alpha -\beta) *  L_{dis\_serv}, \nonumber
\end{equation}

\noindent where $\alpha$ and $\beta$ are scalars which regularize the influence of the terms. $L_{dis\_cl}$ refers to the distillation loss of past model (teacher 1), defined in the previous section, and $L_{dis\_serv}$ refers to the distillation loss of the server model (teacher 2).

\smallskip

\begin{algorithm}[h!]
		\caption{FLwF-2.}
		\label{alg:fcl}
		\begin{algorithmic}[1]
            \Procedure{Server executes:}{}
			\State 
			initialize server model by $\theta_0$
			\For {round  $r = 1,2,...,R$} 
			\State m = 0
			\For {client $k = 1, ..., K$ \textbf{in parallel}} 
			
			\State $\theta_{r}^k \leftarrow$ ClientUpdate$( k, r, \theta_{r-1},\theta_{r-1}^k )$ 
			
			\State $m_k = |D_{kr}|$ \hspace{10mm} // \textit{Size of a dataset} $D_{kr}$
			\State $m += m_k$
			
			\EndFor
			\State  $\theta_{r} \leftarrow \sum\limits_{k=1}^{K} \frac{m_{k}}{m} \theta_{r}^{k}$
			\EndFor
			
			\EndProcedure
			
			\smallskip

            \Function{ClientUpdate}{$k, r, \theta_{r-1},\theta_{r-1}^k$}:    
			
			\State $\mathcal{B} \leftarrow\left(\right.$split $D_{kr}$ into batches of size $\left.B\right)$
			\State $\theta^k_r = \theta_{r-1} $
			\For {each local epoch $i$ from $1$ to $E$ }
	        \For {batch $b \in \mathcal{B} $}
	        \State $\theta^k_r \leftarrow \theta^k_r-\eta \nabla L_{FLwF-2}(\theta, \theta_{r-1},\theta_{r-1}^k; b)$
	        \EndFor
	        \EndFor
	        \State return $\theta^k_r$ to the server
			\EndFunction
		\end{algorithmic}
\end{algorithm}

\vspace{-0.35cm}

\medskip

The proposed solution helps to transfer knowledge from the server  and decreases the forgetting of previously learnt tasks.  It is a regularization-based approach as it prevents activation drift while learning new tasks. 

\smallskip

The pseudo-code of FLwF-2 is presented in Algorithm \ref{alg:fcl}, where the loss function is computed on a batch $b$, $B$ is the local mini-batch size, $\eta$ is the learning rate for the gradient step in model updating, $L_{FLwF-2}(... ; b)$  is a loss function.

%%%%%%%%%%%%%%%%%%%%%%%%%%%%%%%%%%%%%%%%%%%%%%%%%%%%%%%%%
%%%%%%%%%%%%%%%%%%%%%%%%%%%%%%%%%%%%%%%%%%%%%%%%%%%%%%%%%%
%%%%%%%%%%%%%%%%%%%%%%%%%%%%%%%%%%%%%%%%%%%%%%%%%%%%%%%%%%
%%%%%%%%%%%%%%%%%%%%%%%%%%%%%%%%%%%%%%%%%%%%%%%%%%%%%%%%%%
%%%%%%%%%%%%%%%%%%%%%%%%%%%%%%%%%%%%%%%%%%%%%%%%%%%%%%%%%%
%%%%%%%%%%%%%%%%%%%%%%%%%%%%%%%%%%%%%%%%%%%%%%%%%%%%%%%%%%
%%%%%%%%%%%%%%%%%%%%%%%%%%%%%%%%%%%%%%%%%%%%%%%%%%%%%%%%%%
%%%%%%%%%%%%%%%%%%%%%%%%%%%%%%%%%%%%%%%%%%%%%%%%%%%%%%%%%%
%%%%%%%%%%%%%%%%%%%%%%%%%%%%%%%%%%%%%%%%%%%%%%%%%%%%%%%%%%

\section{Experiments}

In this section, we present the different experiments that have been run to evaluate FLwF-1 and FLwF-2 and the metrics that have been defined for evaluation. 

\subsection{Metrics}

\subsubsection{Notations}

$a_{t,d}^{kr}$ is the accuracy of the model trained during communication round $r$ on a task $d$ after learning task $t$ ($d \leqslant t$) for client $k$. To compute $a_{t,d}^{kr}$, we take all the examples of classes of task $d$ present in the test set, calculate an accuracy of the model, which was trained during communication round $r$, on them after the learning task $t$. An accuracy $a_{0}^{kr}$ is calculated on the whole test set after communication round $r$ for client $k$ or the server.

\smallskip
\smallskip

\subsubsection{Metrics to measure FL effectiveness}

To evaluate the effect of the FL, the general and personal accuracy were computed. General accuracy is calculated for the observed client, the generalized client and the server. It is defined as follow:

    \vspace{-0.29cm}
    \begin{equation}
    \label{eq:gen}
    A_{gen}^k = \frac{1}{R} \sum_{r = 1}^R a_{0}^{kr}. \nonumber
    \end{equation}
    \vspace{-0.25cm}
    
Personal accuracy is defined as it follows: 

\begin{equation}
    \label{eq:per}
    A_{per}^k = \frac{1}{R} \sum_{r = 1}^R a_{per}^{kr}, \nonumber
\end{equation}

\noindent where an accuracy  $a_{per}^{kr}$ is calculated on classes which  were already learnt by a client $k$ during the rounds $1,...,r$. 

\subsubsection{Metrics to measure Forgetting}

To evaluate how a model forgets tasks already learned, we compute an average accuracy and a forgetting measure \cite{class-inc}. 

The average accuracy of task $t$ ($A_t^k$) 
for client $k$ is calculated by averaging the accuracy over all the rounds of the sequences of tasks till $t$:

 \begin{equation}
    \label{eq:avgacc}
    A_{t}^k = \frac{1}{t} \sum_{d = 1}^{t} a_{t,d}^{k}, \hspace{7mm} a_{t,d}^{k} = \frac{1}{r_k^t} \sum_{r'= R_k^{t-1}}^{R_k^{t-1} + r_k^t} a_{t,d}^{kr'}, \nonumber
\end{equation}
\vspace{-0.3cm}

\noindent where $R_k^t = \sum_{d=1}^{t}r^{d}_{k}$.  

\smallskip
\smallskip

Forgetting ($f_{t,d}^k$) shows how the model forgets the knowledge about task $d$ after learning task $t$ for client $k$ :
    \vspace{-0.1cm}
    \begin{equation}
    \label{eq:forgetting}
    f_{t,d}^{k}= \max _{i \in\{d, \ldots, t-1\}}\{a_{i,d}^{k}\}-a_{t,d}^{k}. \nonumber
\end{equation}
\vspace{-0.3cm}

It can be averaged as  $F_{t}^k=\frac{1}{t-1} \sum_{d=1}^{t-1} f_{t,d}^{k}$. The higher $F_{t}^k$ is, the more a model forgets. Hence, $F_{t}^k$ must tend to zero.

\subsection{Scenario}

To evaluate the FLwF-1 and FLwF-2 algorithms, we used the scenario defined in section III. It is based on the notions of observed client and of generalized client. The observed client, called Client 1, carries out specific activities in a way that is favorable to catastrophic forgetting. The generalized client allows to normalize the behavior of many clients (the \textit{weight} of this client is taken into account when aggregating the models). This client receives tasks, between each FL round, which are built randomly using the UCI dataset.

In the defined scenario, Client 1 learns two activities, each of them during four rounds with different data. These activities are successively \emph{Walking upstairs (1)} during the first four rounds and \emph{Walking downstairs (2)}  in  the  next  four  rounds. 

We have conducted several experiments that are presented here after. In order to allow comparison, we also implemented fine tuning in the context of Federated Learning. Fine-tuning performs training with a cross-entropy loss and with no knowledge of the tasks treated in the previous communication rounds. This baseline is commonly used in Continual Learning works to demonstrate the problem of catastrophic forgetting. 

For all the experiments, we computed the general and personal accuracy of the clients and of the server. We also computed the average accuracy and forgetting for the observed clients, as defined before.

For {FLwF\-1} and {FLwF-2}, we used a temperature  scaling  parameter $T=2$ as commonly used in other experiments \cite{bic}, \cite{eeil}, \cite{class-inc}. By using the grid search algorithm, we found that for {FLwF-1} $\alpha = 0.001$, and for {FLwF-2}  $\alpha = 0.001$ and $\beta = 0.7$ show the best performances.

%%%%%%%%%%%%%%%%%%%%%%%%%%%%%%%%%%%%
%%%%%%%%%%%%%%%%%%%%%%%%%%%%%%%%%%%%

\subsection{Experiment 1: Distillation for all}

The purpose of this first experiment is to use distillation-based techniques for all the clients, whether they have to deal with unbalanced tasks or not. Table~\ref{tab:models2} and \ref{tab:models3} show how FLwF-1 and FLwF-2 (blue row) are positioned with respect to fine tuning. The best scores are in bold. Regarding general accuracy, it appears that the distillation-based approaches outperform fine-tuning for Client $1$. Knowledge transfer from the previous client model is happening as expected. Results are slightly better with  FLwF-2, which indicates that some knowledge transferred from the server is also preserved. In contrast, general accuracy of the generalized client and of the server are much better with fine tuning. We believe that, for well balanced tasks, distillation-based approaches are less adapted than fine tuning. And since we are in a FL process, these bad results of the generalized client may be propagated on client $1$.  

Forgetting results are provided on Table~\ref{tab:models3}. It clearly appears that despite a close average accuracy  for all three approaches, fine-tuning exhibits a massive catastrophic forgetting behaviour. However, the distillation-based approaches, FLwF-1 and 2 are able to mitigate the forgetting effect. 

\medskip

\definecolor{LightCyan}{rgb}{0.8,1,1}

\begin{table}[!ht]
  \begin{center}
   \begin{tabular}{|c|c|c|c|c|}
      \hline
      \textbf{Method} & \textbf{$A_{gen}^1$} & \textbf{$A_{gen}^g$} & \textbf{$A_{gen}^{server}$} & \textbf{$A_{per}^1$} \\ [0.05cm]
   \hline
       Fine Tuning in FL & 0.478 & \textbf{0.794} & \textbf{0.714} & \textbf{0.750}\\ [0.05cm]
      \hline 
      \rowcolor{LightCyan}
      FLwF-1 & 0.629 & 0.673 & 0.671 & 0.628 \\
      \rowcolor{LightCyan}
      FLwF-2 &  \textbf{0.655} & 0.679 &  0.680 & 0.629\\ [0.05cm]
      \hline 
\end{tabular}

\medskip

  \caption{\textbf{FL metrics}: general accuracy ($A_{gen}^k$) and personal accuracy ($A_{per}^k$) for Client 1 ($k=1$), Generalized client ($k=g$) and Server ($k=server$).}
    \label{tab:models2}
  \end{center}
\end{table}

\vspace{-0.7cm}

\begin{table}[!ht]
  \begin{center}
\begin{tabular}{|c|c|c|}
      \hline
      \textbf{Method} & \textbf{$A^1_2$ (Avg Acc.)} $\uparrow$& \textbf{$F^1_2$ (Forgetting)} $\downarrow$\\ [0.1cm]

      \hline
      Fine Tuning in FL & 0.5 & 1  \\ [0.03cm]
      \hline 
      \rowcolor{LightCyan}
      FLwF-1 & 0.535 & 0.595 \\
      \rowcolor{LightCyan}
      FLwF-2 &  \textbf{0.578} & \textbf{0.418} \\ [0.05cm]
      \hline
\end{tabular}

\medskip

    \caption{\textbf{CL metrics}: average accuracy at task $t$ ($A^k_t$) and forgetting ($F_t^k$) for Client 1 ($k=1$) and task 2 ($t=2$). }

    \label{tab:models3}
  \end{center}
\end{table}

\vspace{-0.5cm}

%%%%%%%%%%%%%%%%%%%%%%%%%%%%%%%%%%%%
%%%%%%%%%%%%%%%%%%%%%%%%%%%%%%%%%%%%

\subsection{Experiment 2: Distillation and Fine Tuning}

The purpose of the second experiment is to evaluate a more refined strategy at the client level. Specifically, distillation is only used when the task to be treated is unbalanced, otherwise fine-tuning is preferred. This strategy, more aware of the context and actual data distribution, mainly concerns the generalized client since client $1$ is by definition very unbalanced. Results are provide by Table~\ref{tab:models4} and \ref{tab:models5} where best results are in bold. The results of the previous experiment are recalled (lines FLwF-1 and FLwF-2, in white) to ease comparison. Federated Learning metrics are provided by Table~\ref{tab:models4}. Regarding general accuracy, it clearly appears that our approach outperforms fine-tuning for all clients (both client $1$ and generalized client). It is also apparent that FLwF-2 performs better than FLwF-1 for all metrics. These good results demonstrate that the distribution of classes in the tasks has to be considered and that distillation has to be limited to unbalanced tasks.
Information about forgetting is provided by Table~\ref{tab:models5}. It appears that the smart combination of distillation and fine tuning provides much better results than distillation only. This is particularly evident on the Forgetting metric.

\begin{table}[!ht]
\centering
   \begin{tabular}{|c|c|c|c|c|}
      \hline
      \textbf{Method} & \textbf{$A_{gen}^1$} & \textbf{$A_{gen}^g$} & \textbf{$A_{gen}^{server}$} & \textbf{$A_{per}^1$} \\ [0.1cm]
    \hline
       Fine Tuning in FL & 0.478 & 0.794 & 0.714 & 0.750\\ [0.05cm]      
       \hline 
       FLwF-1 & 0.629 & 0.673 & 0.671 & 0.628 \\
       FLwF-2 & 0.655 & 0.679 &  0.680 & 0.629\\ [0.05cm]
      \hline 
      \rowcolor{LightCyan}
      FLwF-1 \& Fine Tuning & 0.708 & 0.783 & 0.781 & 0.755 \\
      \rowcolor{LightCyan}
      FLwF-2 \& Fine Tuning &  \textbf{0.753} & \textbf{0.802} &  \textbf{0.797} & \textbf{0.798}\\ [0.05cm]
      \hline 

\end{tabular}

\medskip

  \caption{\textbf{FL metrics}: general ($A_{gen}^k$) and personal accuracy ($A_{per}^k$) for Client 1 ($k=1$), Gen. client ($k=g$) and Server ($k=server$).}
    \label{tab:models4}

\end{table}

\vspace{-0.5cm}

\begin{table}[!ht]
\centering
\begin{tabular}{|c|c|c|}
      \hline
      \textbf{Method} & \textbf{$A^1_2$ (Avg Acc.)} $\uparrow$& \textbf{$F^1_2$ (Forgetting)} $\downarrow$\\ [0.1cm]
      
      \hline 
      Fine Tuning in FL & 0.5 & 1  \\ [0.03cm]
      
      \hline 
      FLwF-1 & 0.535 & 0.595 \\
      FLwF-2 & 0.578 & 0.418 \\ [0.05cm]
      \hline 
      \rowcolor{LightCyan}
      FLwF-1 \& Fine Tuning & 0.696 & 0.392 \\
      \rowcolor{LightCyan}
      FLwF-2 \& Fine Tuning &  \textbf{0.760} & \textbf{0.212} \\ [0.05cm]
      \hline
\end{tabular}
\medskip
    \caption{\textbf{CL metrics}: average accuracy at task $t$ ($A^k_t$) and forgetting ($F_t^k$) for Client 1 ($k=1$) and task 2 ($t=2$). }
    \label{tab:models5}
\end{table}

\vspace{0.45cm}

%%%%%%%%%%%%%%%%%%%%%%%%%%%%%%%%%%%%
%%%%%%%%%%%%%%%%%%%%%%%%%%%%%%%%%%%%

\subsection{Scenario 3: Use of stored exemplars}

The third experiment builds on the previous one and, in addition, makes use of exemplars of learned task, a method which has shown good results in Continual Learning. Precisely, we kept in memory of few exemplars of seen tasks for each client. We implemented the following procedure: if a task is new, 10 exemplars of this task are saved in memory; if it has already been learned, the memory is refreshed with new exemplars from this task. We implemented a random sampling strategy commonly used in class-incremental CL that does not require much computational resources \cite{class-inc}. For each client and each round, exemplars are used in the training process like collected data. Distillation loss, which is part of the loss function, is then computed for the exemplars. The results are shown on Table~\ref{tab:models6} and \ref{tab:models7}.

\begin{table}[!ht]
\centering
   \begin{tabular}{|c|c|c|c|c|}
      \hline
      \textbf{Method} & \textbf{$A_{gen}^1$} & \textbf{$A_{gen}^g$} & \textbf{$A_{gen}^{server}$} & \textbf{$A_{per}^1$} \\ [0.1cm]
    
    \hline
       Fine Tuning in FL & 0.478 & 0.794 & 0.714 & 0.750\\ [0.05cm]      \hline 
      
       FLwF-1 & 0.629 & 0.673 & 0.671 & 0.628 \\
       FLwF-2 & 0.655 & 0.679 &  0.680 & 0.629\\ [0.05cm]
      \hline 
      
      FLwF-1 \& Fine Tuning & 0.708 & 0.783 & 0.781 & 0.755 \\
      FLwF-2 \& Fine Tuning &  \textbf{0.753} & \textbf{0.802} &  \textbf{0.797} & \textbf{0.798}\\ [0.05cm]
      \hline 
      \rowcolor{LightCyan}
      FLwF-1 \& Fine Tuning \& ex &  0.705 & 0.789 & 0.779 & 0.748 \\
      \rowcolor{LightCyan}
      FLwF-2 \& Fine Tuning \& ex &  0.750  & 0.781 & 0.778 & 0.755 \\ [0.05cm]
      
      \hline 

\end{tabular}

\medskip
    
      \caption{\textbf{FL metrics}: general ($A_{gen}^k$) and personal accuracy ($A_{per}^k$) for Client 1 ($k=1$), Gen. client ($k=g$) and Server ($k=server$).}
    
    \label{tab:models6}
\end{table}

\begin{table}[!ht]
\centering
\begin{tabular}{|c|c|c|}
      \hline
      \textbf{Method} & \textbf{$A^1_2$ (Avg Acc.)} $\uparrow$& \textbf{$F^1_2$ (Forgetting)} $\downarrow$\\ [0.1cm]
      
      \hline 
      Fine Tuning in FL & 0.5 & 1  \\ [0.03cm]
      
      \hline 
      FLwF-1 & 0.535 & 0.595 \\
      FLwF-2 & 0.578 & 0.418 \\ [0.05cm]
      \hline 

      FLwF-1 \& Fine Tuning & 0.696 & 0.392 \\
      FLwF-2 \& Fine Tuning &  \textbf{0.760} & 0.212 \\ [0.05cm]
      \hline

      \rowcolor{LightCyan}
      FLwF-1 \& Fine Tuning \& ex &  0.678 & 0.407\\
      \rowcolor{LightCyan}
      FLwF-2 \& Fine Tuning \& ex & 0.746 & \textbf{0.12}  \\ [0.05cm]
      \hline
      
\end{tabular}

\medskip

    \caption{\textbf{CL metrics}: average accuracy at task $t$ ($A^k_t$) and forgetting ($F_t^k$) for Client 1 ($k=1$) and task 2 ($t=2$). }

    \label{tab:models7}
\end{table}

\begin{figure*}[!ht]
\centering
\begin{minipage}{.6\textwidth}
  \centering

 \includegraphics[width=1.1 \linewidth]{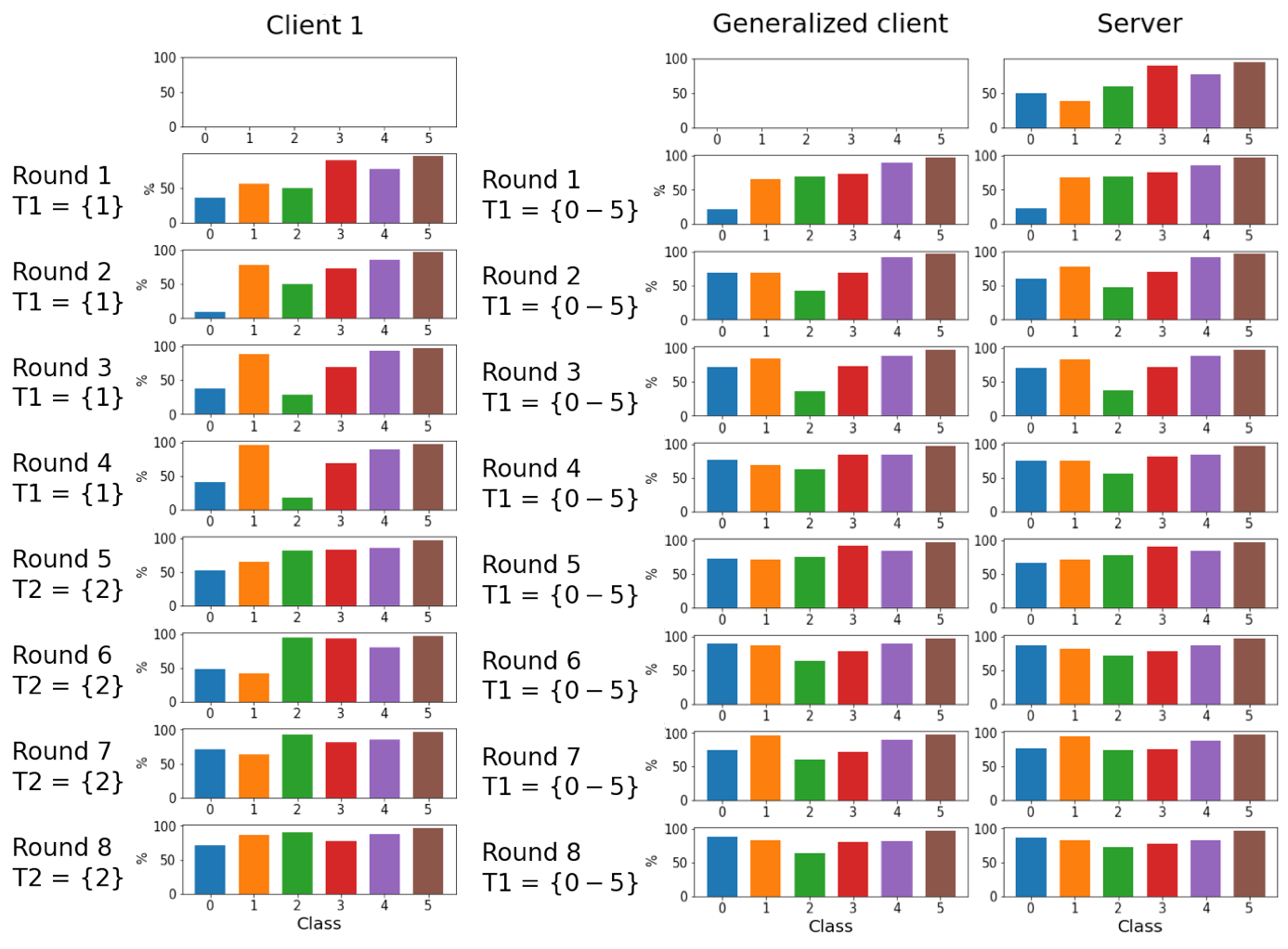}

  \label{fig:test1}
\end{minipage}%

\smallskip

\caption{Results obtained with FLwF-2\&Fine-Tuning for the scenarios of Client~1\label{fig:comparison}}
\end{figure*}

\subsection{Graphical visualization of the clients behaviour}

Figure~\ref{fig:comparison} provides a different perspective on the results. It shows how accuracy %, that is the percentage of correctly classified examples from the test set, 
evolves after each of the eight rounds. This figure highlights several important points. On the left part of the figure (Client~1), we can see that the forgetting phenomenon is greatly reduced compared to figure 3, which is based on the same scenario. Moving from Activity~1 (\emph{Walking upstairs}) to Activity~2 (\emph{Walking downstairs}) does not result in a catastrophic forgetting effect. We can also see very well, with activity 0 (\emph{Walking}) for example, that the transfer of knowledge from the generalized client to Client~1 is done well and is preserved thanks to distillation.

\medskip
\medskip

\section{Conclusion}

Dealing with catastrophic forgetting is a necessary step for providing lifelong-learning in smart collection of objects. In this paper, we present two versions of a Federated Continual Learning (FCL) approach (FLwF-1 and FLwF-2) to deal with catastrophic forgetting in pervasive computing. This approach relies on a distillation technique to transfer generic knowledge to specialized models. Evaluation has shown that an FCL approach that combines context-aware distillation and fine-tuning greatly reduces the catastrophic forgetting effect. 

Precisely, we proposed distillation-based approaches, FLwF-1 and FLwF-2, based respectively on one and two teachers. It clearly appeared that the use of two teachers, one coming from the client and one coming from the server prevents catastrophic forgetting more effectively. It allows increasing general knowledge of clients while preserving its private past knowledge. Furthermore, our solution does not require higher computational and storage resources than Federated Learning since it uses models that are stored anyway  at each Federated Learning round.

\section{Acknowledgements}
This work was partially supported by MIAI@Grenoble-Alpes (ANR-19-P3IA-0003).

\bibliographystyle{plain}

{\footnotesize
\bibliography{smartcomp}}

\end{document}